\documentclass[conference]{IEEEtran}

\usepackage[utf8]{inputenc}
\usepackage{cite}
\usepackage{amsmath,amssymb}
\usepackage{graphicx}
\usepackage{lipsum} 
\usepackage[hidelinks]{hyperref} 

\providecommand{\keywords}[1]{\textbf{Keywords:} #1}

\title{Financial Decision Making using Reinforcement Learning with Dirichlet Priors and Quantum-Inspired Genetic Optimization}

\author{
\IEEEauthorblockN{Prasun Nandy, Debjit Dhar\IEEEauthorrefmark{1}, Rik Das}
Data \& Analytics, PwC Inida
\IEEEauthorblockA{\IEEEauthorrefmark{1}Correspondence: debjitd.cse.ug@jadavpuruniversity.in}

}

\begin{document}

\maketitle

\begin{abstract}
Traditional budget allocation models struggle with the stochastic and nonlinear nature of real-world financial data. This study proposes a hybrid reinforcement learning (RL) framework for dynamic budget allocation, enhanced with Dirichlet-inspired stochasticity and quantum mutation-based genetic optimization. Using Apple Inc.’s quarterly financial data (2009–2025), the RL agent learns to allocate budgets between R\&D and SG\&A to maximize profitability while adhering to historical spending patterns, with L2 penalties discouraging unrealistic deviations. A Dirichlet distribution governs state evolution to simulate shifting financial contexts. To escape local minima and improve generalization, the trained policy is refined using genetic algorithms with quantum mutation via parameterized qubit rotation circuits. Generation-wise rewards and penalties are logged to visualize convergence and policy behavior. On unseen fiscal data, the model achieves high alignment with actual allocations (cosine similarity 0.9990, KL divergence 0.0023), demonstrating the promise of combining deep RL, stochastic modeling, and quantum-inspired heuristics for adaptive enterprise budgeting.
\end{abstract}

\section{Introduction}
Strategic budget allocation is a cornerstone of corporate financial planning \cite{10.1371/journal.pone.0288627}, especially in technology-driven firms such as Apple Inc., where the balance between innovation and operational efficiency can define long-term competitiveness. Among the most impactful financial decisions lies the allocation of limited resources between Research and Development (R\&D)—which drives future innovation—and Selling, General, and Administrative (SG\&A) expenses—responsible for sustaining ongoing business operations and market presence \cite{10.1093/icc/11.3.529}.

Traditional budgeting approaches rely heavily on historical trends, executive judgment, and static rules-based models. While these methods offer interpretability, they are inherently limited in their ability to respond dynamically to market volatility, macroeconomic shocks, or the long-term consequences of prior decisions. Moreover, they often ignore the sequential and uncertain nature of financial planning, resulting in reactive allocations that may not align with strategic goals in a multi-quarter horizon.
Manual and rule-based budget allocation methods \cite{CAO2025247} often struggle with integrating large volumes of longitudinal data, adapting to dynamic economic environments, and accounting for competing objectives across departments. Human decision-makers are also prone to cognitive biases, short-termism, and political pressures, all of which can lead to suboptimal or inconsistent allocations. Additionally, such methods tend to be reactive rather than proactive, making it difficult to anticipate the financial consequences of budgetary decisions over multiple fiscal cycles. These challenges are exacerbated in global firms operating in fast-changing industries, where even small misalignments in resource allocation can cascade into lost opportunities, operational inefficiencies, or misdirected innovation efforts.

\noindent Amid these limitations, artificial intelligence (AI) has 
emerged as a promising avenue for augmenting financial decision making. \cite{10895709} Machine learning models can uncover complex patterns in financial data, forecast trends, and assist in scenario analysis. In particular, AI techniques can help shift budgeting from a backward-looking activity to a forward-looking optimization problem. However, most traditional AI applications in finance—especially supervised learning—are limited in their ability to handle sequential decision-making and feedback-driven environments. They typically predict outcomes based on fixed inputs and do not account for the evolving nature of financial systems or the long-term impact of decisions.

Reinforcement learning (RL) \cite{ghasemi2024introductionreinforcementlearning,han2020actorcriticreinforcementlearningcontrol}, by contrast, offers a more natural fit for complex budget planning tasks. Unlike supervised learning, RL agents learn to make decisions through interaction with their environment, receiving feedback in the form of rewards or penalties. This makes RL especially suitable for modeling financial systems where actions influence future states and where optimality must be measured over extended time horizons. In budget allocation scenarios, RL can account for delayed returns on investment, dynamic constraints, and evolving organizational priorities—all of which are difficult to model explicitly in traditional systems \cite{712192, bai2024reviewreinforcementlearningfinancial}. As financial environments grow more uncertain and multidimensional, RL’s ability to adaptively learn policies that balance multiple objectives over time becomes increasingly valuable.

Despite its potential, applying RL to real-world corporate finance presents non-trivial challenges. Financial environments are often noisy, partially observable, and constrained by domain-specific rules such as regulatory guidelines, internal controls, and organizational inertia. Moreover, decision-makers must be able to interpret and trust AI-driven policies—making transparency and constraint adherence just as important as raw performance. To be truly effective, AI-driven budgeting tools must not only optimize financial outcomes but also align with realistic business practices and risk management protocols.
To address these limitations, this study presents a novel AI-driven framework that leverages deep reinforcement learning (RL) for dynamic budget allocation. The approach is grounded in a custom-built simulation environment using real-world quarterly financial data from Apple Inc.\ (2009–2025) \cite{macrotrends_aapl}, where an RL agent learns to allocate budgets between R\&D and SG\&A in a way that maximizes net income while staying close to historically realistic ratios. The environment incorporates Dirichlet-based Bayesian updating to simulate evolving budget priors, ensuring a realistic feedback loop that mimics how organizations adjust financial priorities over time.

The learning agent is trained using the Twin Delayed Deep Deterministic Policy Gradient (TD3) algorithm from the Stable-Baselines3 library, with exploration guided by Gaussian action noise. The reward function explicitly balances profitability against an L2 penalty for deviating from real-world allocation patterns—encouraging both fiscal performance and interpretability. After training, the agent undergoes policy fine-tuning via a quantum-inspired genetic optimizer, where PennyLane-based quantum circuits introduce noise-driven mutations to help escape local optima and promote generalization in out-of-sample scenarios.

The objectives of this work are as follows:
\begin{itemize}
    \item To investigate the limitations of traditional and rule-based budget allocation methods in dynamic and uncertain financial environments, particularly in technology-driven firms where the trade-off between innovation and operational efficiency is critical.
    
    \item To develop a deep reinforcement learning (RL)-based framework for dynamic corporate budget allocation that adaptively learns optimal policies for allocating funds between R\&D and SG\&A expenditures over multiple financial quarters.
    
    \item To simulate a realistic financial decision-making environment using real-world longitudinal data (Apple Inc., 2009--2025), incorporating Bayesian updating to reflect evolving budgetary priorities and constraints aligned with business practices.
    
    \item To enhance model performance and generalization by integrating a quantum-inspired genetic optimizer for fine-tuning RL policies, thus addressing challenges related to local optima, interpretability, and policy robustness in real-world financial planning scenarios.
\end{itemize}

The integrated approach followed in this research addresses key gaps in current financial decision systems. It models sequential dependencies, learns from real data, incorporates business realism through probabilistic priors and domain-aligned rewards, and explores cutting-edge optimization via quantum-inspired techniques~\cite{10823421}. By demonstrating this methodology in a real-world financial context, the study contributes to the emerging literature on AI-augmented corporate finance, offering a blueprint for intelligent, adaptive, and interpretable budgeting systems.

\section{Related Work}
Traditional financial modeling and budget optimization techniques—including linear programming, Markov decision processes (MDPs), and econometric time series models—have long served as foundational tools in quantitative finance. While effective in well-structured, low-dimensional settings, these approaches rely heavily on assumptions such as stationarity, convexity, and complete observability. These constraints render them insufficient for modern financial systems characterized by nonlinear interactions, multi-objective trade-offs, and evolving uncertainty.

To address these limitations, Reinforcement Learning (RL) has gained prominence as a sequential decision-making paradigm that learns optimal policies through interaction and feedback \cite{bai2024reviewreinforcementlearningfinancial}. Classical RL methods, however, such as tabular Q-learning or policy iteration, are not scalable to high-dimensional, continuous-action spaces relevant in corporate budgeting.

Recent advancements in Deep Reinforcement Learning (DRL)—notably DDPG, TD3, PPO, and SAC—have significantly enhanced the applicability of RL to real-world financial domains by leveraging neural function approximators  \cite{mnih2013playingatarideepreinforcement, a16010023}. These models can capture complex dynamics \cite{nakayama2023causalinferenceinvestmentconstraints} and have demonstrated success across asset allocation, trading, and credit scoring.

For instance, Huang et al. \cite{huang2025deepreinforcementlearningframework} introduced a DRL actor-critic model optimized via the Sharpe ratio using image-based state representations, showcasing improved convergence and robust portfolio behavior. Similarly, Espiga-Fernandez et al. \cite{a17120570} conducted a large-scale evaluation of DRL agents, highlighting how feature encoding and reward shaping critically influence long-term returns.

A major milestone in incorporating investor preference and market frictions came from Jiang et al. \cite{JIANG2024101016}, who used a TD3 framework with mean-variance optimization under transaction costs, leading to robust financial policy learning under noisy and volatile market regimes.

Despite these gains, real-world financial systems often involve delayed effects, regime shifts, and partial observability—conditions where classical DRL struggles. To address these issues, Bayesian Reinforcement Learning (BRL) has been proposed as a way to model and act under explicit uncertainty. In BRL, policies are updated using probabilistic belief models rather than point estimates. Vlassis et al. \cite{Vlassis2012} provide a comprehensive overview of BRL where priors encode domain knowledge, posterior updates adapt policies dynamically, and exploration is treated as probabilistic inference. These ideas are further reinforced by Kang et al. \cite{KANG2024107924}, who reinterpret the Rescorla-Wagner model under a Bayesian lens, tying BRL to neurobiological models of adaptive learning.

Recent work by Roy et al. \cite{roy2025generalizedbayesiandeepreinforcement} advances this direction through Generalized Bayesian DRL, where deep policies are regularized with uncertainty-aware priors and posterior adaptation improves generalization in dynamic environments. The convergence and complexity trade-offs of such Bayesian methods are well analyzed in \cite{Ghavamzadeh_2015}, making a case for scalable, sample-efficient probabilistic reinforcement frameworks.

To further support robust long-term learning, hybrid methods have been gaining ground. Preil and Maier \cite{preil2023geneticmultiarmedbanditsreinforcement} proposed the GMAB framework, combining genetic algorithms (GAs) with multi-armed bandits to enhance policy exploration in discrete budget scenarios. Their model captures the global search strengths of evolutionary algorithms and the local refinement of bandits, aligning with the philosophy of multi-resolution search.

Earlier foundational work by Allen and Karjalainen \cite{ALLEN1999245} showed that GAs could effectively uncover technical trading rules, emphasizing their suitability in rule discovery and non-convex optimization. However, standard GAs often converge prematurely or get stuck in local minima. To mitigate this, quantum-inspired variants have been explored.

Narayanan and Moore \cite{narayanan1996quantum} demonstrated that Quantum Genetic Algorithms (QGAs), which exploit qubit-based representations and quantum gate operations, outperform classical GAs in convergence speed and exploration balance. While promising, QGA applications to finance remain underexplored, especially in policy search and hyperparameter optimization for RL systems.

Recent ensemble-based methods like Naik and Albuquerque's hybrid model \cite{article} combine XGBoost, ANNs, and Squirrel-Whale inspired optimization for predictive modeling and asset allocation. These works demonstrate that blending neural, tree-based, and bio-inspired heuristics can lead to stronger generalization and adaptability in non-stationary settings.

Despite these innovations, few existing models integrate DRL, Bayesian inference and quantum-enhanced evolutionary strategies within a cohesive architecture. Most efforts remain siloed—either focusing on RL policy improvement or heuristic search, without leveraging their combined potential.

While each of the aforementioned techniques—Deep Reinforcement Learning (DRL), Bayesian Reinforcement Learning (BRL), Genetic Algorithms (GAs), and ensemble methods—offers unique advantages, they have mostly been explored in isolation. Few existing frameworks attempt to unify these approaches into a cohesive, simulation-driven architecture for financial decision-making.

To address this gap, we propose a modular, multi-agent architecture that integrates:

\begin{itemize}
    \item TD3-based Deep Reinforcement Learning, for learning adaptive policies in continuous financial domains,
    \item Dirichlet-based state evolution, to introduce controlled stochasticity in budget transitions,
    
    \item Quantum Genetic Algorithms (QGAs), to enhance policy diversity and avoid convergence to suboptimal strategies.
\end{itemize}

Built with modular code, our framework enables flexible policy tuning, interpretable feedback signals, and simulation-based testing under real-world financial constraints. By merging learning, control, and evolutionary computation into a unified system, we aim to deliver a more resilient, interpretable, and adaptive solution to financial planning and budget optimization.

\section{Methodology}

\begin{figure*}[htbp]
  \centering
  \includegraphics[width=\textwidth, keepaspectratio=true]{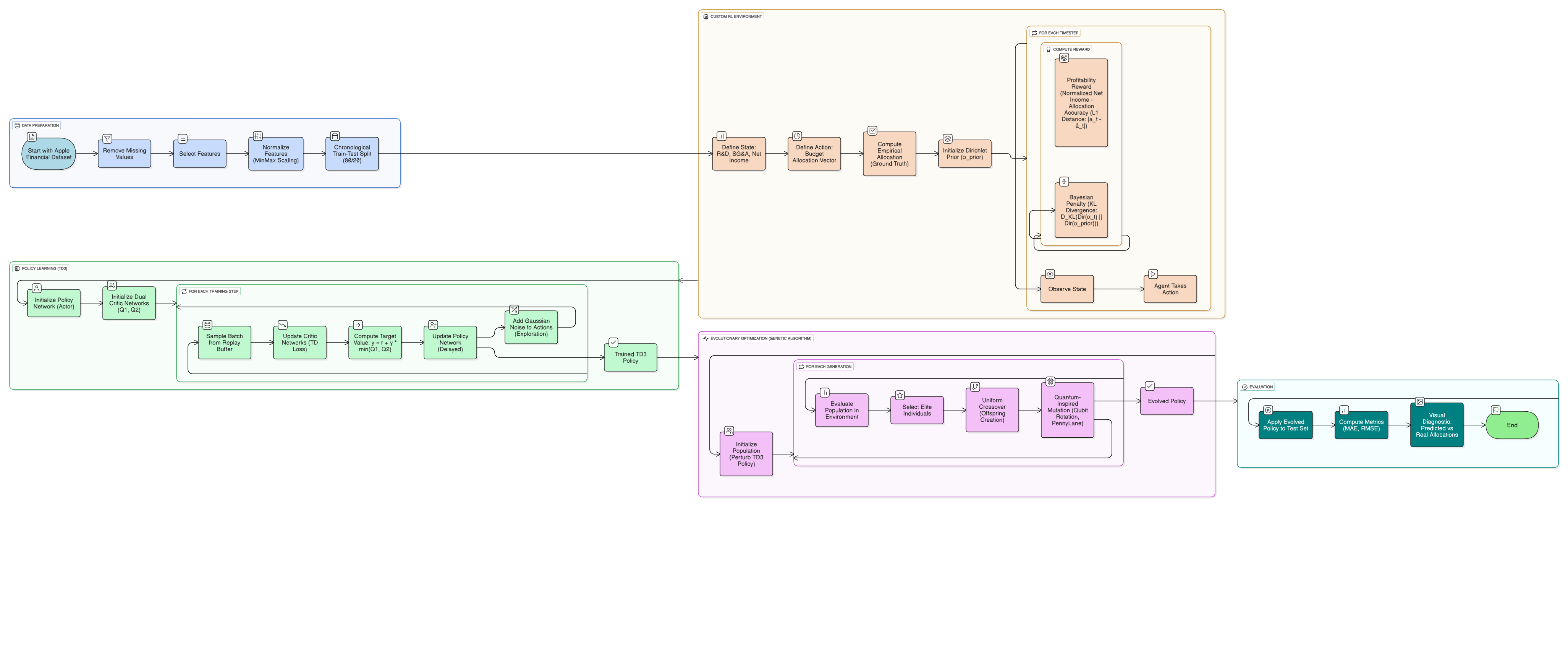}
  \caption{Methodology Pipeline}
  \label{fig:methodology}
\end{figure*}

The proposed methodology outlines a robust and modular framework for financial decision-making under uncertainty, leveraging a blend of reinforcement learning, Bayesian inference, and evolutionary optimization. The pipeline begins with the preprocessing of a real-world Apple financial dataset, where key features such as R\&D expenses, SG\&A expenses, and net income are selected and normalized using MinMax scaling. A chronological split ensures that the temporal structure of the financial data is preserved for training and evaluation, thus avoiding data leakage.
At the heart of the methodology is a custom Gym-compatible environment designed to simulate financial allocation decisions. This environment formalizes budget allocation as a continuous action-space problem, where actions represent the proportion of budget allocated to R\&D and SG\&A. The remaining budget is inferred as the residual. A critical innovation here is the use of Dirichlet priors to encode uncertainty in the budget distribution. With each environment step, the system updates a Dirichlet belief vector—modeled as a prior over three budget categories—based on the agent’s actions, resulting in a Bayesian belief update. This stochastic yet principled approach allows the agent to iteratively refine its understanding of plausible budget allocations over time, based on empirical data.
The reward function integrates both profitability and allocation accuracy. Profit-based reward is calcu- lated as a normalized difference between net income and total expenses. To enforce realism and penalize divergence from actual financial behavior, the framework introduces a regularization penalty using L2 dis- tance between the agent’s action and real-world allocation ratios. This dual-objective reward guides the policy towards both fiscal efficiency and behavioral fidelity.
Policy learning is driven by the Twin Delayed Deep Deterministic Policy Gradient (TD3) algorithm, which is chosen for its effectiveness in continuous control tasks and its ability to reduce overestimation bias via dual critic networks. A Gaussian noise process is applied during training to promote exploration.
To enhance policy generalizability and explore beyond local optima, the trained TD3 model undergoes evolutionary fine-tuning via a genetic algorithm. The GA incorporates population-based search with elite selection, uniform crossover, and a quantum-inspired mutation operator, which injects stochastic perturba- tions based on qubit simulations using PennyLane. This hybridization balances exploitation from gradient descent with exploration from global search.
The trained and evolved policy is finally evaluated on a held-out test set, and its allocation predictions are compared against actual financial ratios using standard metrics such as MAE and RMSE. The pipeline concludes with a visual diagnostic comparing predicted and real allocations, validating that the learned policy not only maximizes reward but also mimics realistic financial strategies under budgetary uncertainty. The methodology flowdiagram is shown in \autoref{fig:methodology}.

\subsection{Dataset and Preprocessing}

We employ a quarterly financial dataset of Apple Inc., covering the period from 2009 to 2025, and encompassing three principal financial indicators:
\begin{itemize}
    \item Research and Development (R\&D) Expenses
    \item Selling, General \& Administrative (SG\&A) Expenses
    \item Net Income
\end{itemize}

\noindent Preprocessing steps included the elimination of records with missing values to ensure data integrity. Subsequently, feature normalization was applied to rescale the variables to the [0, 1] interval using Min-Max scaling, thereby facilitating uniform feature contribution during downstream modeling tasks:

\begin{equation}
x_{\text{scaled}} = \frac{x - x_{\min}}{x_{\max} - x_{\min}}
\tag{1}
\end{equation}
Here, \( x \) is the original value, \( x_{\min} \) and \( x_{\max} \) are the minimum and maximum values in the dataset respectively, and \( x_{\text{scaled}} \) is the normalized value rescaled to the [0, 1] range.

\noindent Min-max normalization is applied to rescale all features—namely R\&D, SG\&A, and Net Income—to a uniform domain $[0, 1]$. This standardization ensures that gradient-based learning algorithms such as TD3 are not skewed by variable magnitude differences, improving convergence behavior. Moreover, bounded action spaces in reinforcement learning benefit from normalized inputs, facilitating stable policy learning in environments with continuous action domains.

\noindent Temporal dependencies were preserved through chronological splitting: 80\% for training and 20\% for evaluation.

A custom OpenAI Gym \cite{brockman2016openaigym} environment was constructed to simulate financial decision-making with probabilistic feedback and Bayesian updates. At each timestep $t$, the agent observes:

\begin{equation}
s_t = \left[ \text{R\&D}_t, \text{SG\&A}_t, \text{NetIncome}_t \right]
\tag{2}
\end{equation}

\begin{equation}
a_t = \left[ a_t^{(1)}, a_t^{(2)} \right], \quad \text{where } a_t^{(1)} + a_t^{(2)} = 1
\tag{3}
\end{equation}

\begin{equation}
\hat{a}_t = \left[ \frac{\text{R\&D}_{t+1}}{\text{R\&D}_{t+1} + \text{SG\&A}_{t+1}}, \frac{\text{SG\&A}_{t+1}}{\text{R\&D}_{t+1} + \text{SG\&A}_{t+1}} \right]
\tag{4}
\end{equation}

\noindent In Equation (2), $s_t$ is the observed state comprising financial indicators. Equation (3) defines the agent's action $a_t$, a 2-dimensional budget allocation vector over R\&D and SG\&A. Equation (4) computes $\hat{a}_t$, the empirical budget allocation at timestep $t$, normalized to form a probability distribution over categories. This serves as ground-truth feedback for policy evaluation.

\subsection{Reward Function with Bayesian Dirichlet Penalty}

The agent maintains a belief over plausible budget allocations using a Dirichlet distribution. The reward incorporates Bayesian inference to evaluate policy adherence to expected spending behaviors:

\begin{align}
r_t =\ & - \left\| a_t - \hat{a}_t \right\|_1 
- \lambda_1 \left\| a_t - a_{t-1} \right\|_2 \nonumber \\
& - \lambda_2 \cdot D_{\text{KL}}\left( \text{Dir}(\boldsymbol{\alpha}_t) 
\parallel \text{Dir}(\boldsymbol{\alpha}_{\text{prior}}) \right)
\tag{5}
\end{align}

\noindent The first two terms remain the same as in Equation (5) of the original formulation: accuracy and smoothness. The third term introduces a Bayesian penalty that quantifies the divergence between the agent's current belief $\text{Dir}(\boldsymbol{\alpha}_t)$ and its prior $\text{Dir}(\boldsymbol{\alpha}_{\text{prior}})$ via the Kullback–Leibler (KL) divergence.

\paragraph{Belief Update via Dirichlet Posterior.} At each timestep $t$, the agent observes the empirical allocation $\hat{a}_t$ and updates its posterior belief using a Dirichlet conjugate prior:

\begin{equation}
\boldsymbol{\alpha}_t = \boldsymbol{\alpha}_{t-1} + \hat{a}_t \cdot c
\tag{6}
\end{equation}

\noindent where $\boldsymbol{\alpha}_t = [\alpha_t^{(1)}, \alpha_t^{(2)}]$ are the concentration parameters of the Dirichlet posterior, and $c$ is a scaling factor representing the "confidence" or effective sample size of the update. This update accumulates evidence over time, refining the belief distribution over optimal allocations.

\paragraph{KL Divergence Between Dirichlet Distributions.} The KL divergence between the updated posterior and prior Dirichlet distributions is computed as:

\begin{align}
D_{\text{KL}}&\left( \text{Dir}(\boldsymbol{\alpha}_t) \parallel \text{Dir}(\boldsymbol{\alpha}_{\text{prior}}) \right) \nonumber\\
&= \log \frac{\Gamma(\sum_j \alpha_t^{(j)})}{\Gamma(\sum_j \alpha_{\text{prior}}^{(j)})}
- \sum_j \log \frac{\Gamma(\alpha_t^{(j)})}{\Gamma(\alpha_{\text{prior}}^{(j)})} \nonumber\\
&\quad + \sum_j (\alpha_t^{(j)} - \alpha_{\text{prior}}^{(j)}) 
\left( \psi(\alpha_t^{(j)}) - \psi\left( \sum_k \alpha_t^{(k)} \right) \right)
\tag{7}
\end{align}

\noindent where $\Gamma(\cdot)$ is the gamma function and $\psi(\cdot)$ is the digamma function. This divergence acts as a regularization term penalizing belief shifts inconsistent with the prior.

\paragraph{Prior Calibration.} The prior $\boldsymbol{\alpha}_{\text{prior}}$ encodes domain knowledge regarding historically stable budget allocations. For instance:

\begin{equation}
\boldsymbol{\alpha}_{\text{prior}} = \left[ \alpha_0^{(1)}, \alpha_0^{(2)} \right] = [5.0, 3.0]
\tag{8}
\end{equation}

\noindent reflects a prior belief that R\&D is typically weighted more heavily than SG\&A. The learning process gradually refines this belief based on observed empirical feedback $\hat{a}_t$.

\paragraph{Summary of Reward Components.} The revised reward balances three objectives:

\begin{itemize}
    \item \textbf{Empirical Accuracy} ($\|\mathbf{a}_t - \hat{\mathbf{a}}_t\|_1$): Measures deviation from the observed real-world allocation at $t+1$.
    \item \textbf{Temporal Smoothness} ($\|\mathbf{a}_t - \mathbf{a}_{t-1}\|_2$): Prevents abrupt shifts in budget decisions.
    \item \textbf{Belief Coherence} ($D_{\text{KL}}$): Penalizes excessive divergence from learned priors, enforcing epistemic consistency.
\end{itemize}

\noindent Unlike deterministic SPC rule-based penalization, this formulation allows the agent to operate under epistemic uncertainty and to evolve its belief distribution based on empirical feedback through Dirichlet posterior updates. This Bayesian approach is better suited to real-world financial environments where underlying allocation patterns are stochastic and prior knowledge can guide exploration.

\subsection{TD3-Based Policy Learning}

We use the Twin Delayed Deep Deterministic Policy Gradient (TD3) algorithm. The policy network $\pi$ and two critic networks \cite{fujimoto2018addressingfunctionapproximationerror} $Q_1$, $Q_2$ are trained via temporal difference learning:
\begin{equation}
\theta_Q \leftarrow \theta_Q - \alpha \nabla_{\theta_Q} \mathbb{E}_{s,a,r,s'} \left[(Q(s,a) - y)^2 \right]
\tag{9a}
\end{equation}

\begin{equation}
y = r + \gamma \cdot \min(Q_1(s', \pi(s')), Q_2(s', \pi(s')))
\tag{9b}
\end{equation}
\noindent In Equation (9a), $\theta_Q$ represents the parameters of the critic (Q-network), and $\alpha$ is the learning rate. The term $\nabla_{\theta_Q}$ denotes the gradient with respect to $\theta_Q$, and the expectation $\mathbb{E}_{s,a,r,s'}$ is taken over a batch of transitions $(s, a, r, s')$ sampled from the replay buffer. $Q(s,a)$ is the current critic’s estimated Q-value for state $s$ and action $a$, and $y$ is the target value defined in Equation (9b). In Equation (9b), $r$ is the immediate reward, $\gamma$ is the discount factor for future rewards, $s'$ is the next state, and $\pi(s')$ is the next action predicted by the current policy (actor network). $Q_1$ and $Q_2$ are two independently trained critic networks, and the $\min(\cdot)$ operation reduces overestimation bias by taking the lower of the two predicted Q-values at the next timestep.

\noindent Policy updates are delayed relative to critic updates, and noise is added to actions during critic learning for better exploration. The model is then trained for 50,000 timesteps using a MultiLayer Perceptron (MLP) policy.
\cite{pmlr-v9-glorot10a}
\subsection{Genetic Algorithm for Post-TD3 Refinement}

After initial training with TD3, we apply a Genetic Algorithm (GA) to evolve policy weights. This approach enhances policy generalization by searching beyond the local optima reached by gradient-based methods.

\subsection*{Evolutionary Optimization Process}
The genetic algorithm (GA) follows the classical evolutionary paradigm with enhancements designed for improved exploration and adaptability in financial control tasks:

\begin{itemize}
    \item \textbf{Population Initialization:} A population is initialized by perturbing the base TD3 policy parameters with Gaussian noise:
    
    \begin{equation}
    \theta_i = \theta_{\text{base}} + \epsilon_i, \quad \epsilon_i \sim \mathcal{N}(0, \sigma^2)
    \tag{10}
    \end{equation}
    
    Here, $\theta_i$ denotes the parameters of the $i$-th individual in the population, $\theta_{\text{base}}$ is the base TD3 actor policy, and $\epsilon_i$ is the perturbation noise sampled from a zero-mean Gaussian distribution with variance $\sigma^2$.

    \item \textbf{Evaluation and Selection:} Each policy is evaluated in the budget allocation environment using cumulative rewards as fitness. The top-$k$ elite individuals (determined by the elite fraction) are selected for reproduction.

    \item \textbf{Crossover:} New offspring are generated using uniform crossover across gene indices:
    
    \begin{equation}
    \theta^{(j)}_{\text{new}} = 
    \begin{cases}
        \theta^{(j)}_a & \text{with probability } 0.5 \\
        \theta^{(j)}_b & \text{otherwise}
    \end{cases}
    \tag{11}
    \end{equation}
    
    In this equation, $\theta^{(j)}_{\text{new}}$ represents the $j$-th gene (parameter) of the new offspring. It is selected from either parent $a$ or parent $b$ with equal probability, enabling genetic diversity in offspring.

    \item \textbf{Quantum-Inspired Mutation:} Each policy gene undergoes a mutation inspired by qubit state dynamics:
    
    The gene is represented as a qubit:
    \begin{equation}
    |\psi\rangle = \cos(\theta)|0\rangle + \sin(\theta)|1\rangle
    \tag{12}
    \end{equation}
    
    Here, $|\psi\rangle$ is the quantum state of the gene, where $\theta$ is the angle parameterizing the probability amplitudes of basis states $|0\rangle$ and $|1\rangle$.

    A rotation gate is applied to the qubit:
    \begin{equation}
    R(\Delta\theta) = 
    \begin{bmatrix}
        \cos(\Delta\theta) & -\sin(\Delta\theta) \\
        \sin(\Delta\theta) & \cos(\Delta\theta)
    \end{bmatrix}
    \tag{13}
    \end{equation}

    The matrix $R(\Delta\theta)$ represents a 2D rotation operator, rotating the qubit state by an angle $\Delta\theta$ in Hilbert space.

    The new qubit state becomes:
    \begin{equation}
    |\psi'\rangle = R(\Delta\theta)|\psi\rangle
    \tag{14}
    \end{equation}

    This updates the gene’s quantum state to $|\psi'\rangle$, allowing for a non-local mutation guided by quantum amplitude transformation.

    Upon measurement (collapse), the updated amplitudes probabilistically determine the final value of the gene. This quantum-inspired mutation facilitates richer exploration and has been implemented using the PennyLane library \cite{bergholm2022pennylaneautomaticdifferentiationhybrid}.
    
\end{itemize}

\subsubsection*{Advantages of Quantum-Inspired Mutation}
The mutation mechanism offers the following advantages:
\begin{itemize}
    \item \textbf{Exploration via Superposition:} Modeling genes as quantum states enables maintaining diverse search trajectories in parallel, enhancing coverage of the parameter space.
    \item \textbf{Probabilistic Influence:} The stochasticity introduced by quantum measurement helps escape local optima, a common challenge in financial optimization.
    \item \textbf{Adaptability to Complex Landscapes:} Quantum mutation’s ability to manipulate amplitudes allows the algorithm to navigate multi-modal, noisy reward surfaces effectively.
\end{itemize}

The genetic algorithm was executed for 10 generations with an initial population size of 5, an elite fraction of 0.4, and a mutation rate of 0.1.

\section{Results}

All experiments were executed on the cloud-hosted computing environment featuring an Intel Xeon 2.3 GHz processor with 13 GB of RAM. Preliminary data preprocessing, including imputation, scaling, and traditional supervised learning pipelines, utilized the aforementioned CPU-based infrastructure.

\noindent To facilitate the computational demands of deep reinforcement learning (DRL) and multi-epoch neural training, a GPU-accelerated configuration was employed. This setup leveraged two NVIDIA Tesla T4 GPUs (configured as T4×2), each with 16 GB of GDDR6 memory. These GPUs support mixed-precision arithmetic and CUDA-based acceleration, leading to significant reductions in training time, especially for high-dimensional, parameter-rich models. All implementations were conducted in Python using the PyTorch \cite{paszke2019pytorchimperativestylehighperformance} for deep learning, and Stable-Baselines3 \cite{JMLR:v22:20-1364} for reinforcement learning.

\noindent Reproducibility was ensured through the use of fixed random seeds across NumPy, PyTorch (CPU and CUDA), and the CUDA backend. The results for budget allocation are shown in \autoref{tab:results1} 
\subsection*{Budget Allocation Objective and Configurations}

\begin{figure*}[htbp]
  \centering
  \includegraphics[width=\textwidth]{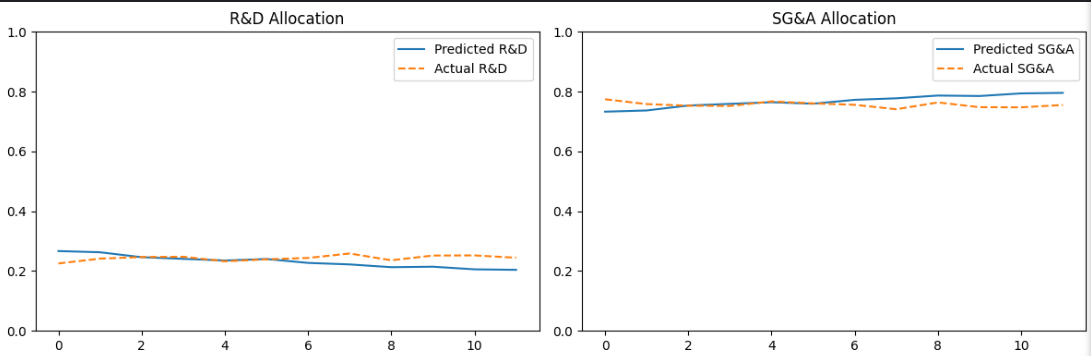}
  \caption{Allocation predictions using RL + SPC + Quantum-Genetic Algorithm}
  \label{fig:results1}
\end{figure*}

The objective of the budget allocation results shown in \autoref{fig:results1} is to evaluate the model's ability to allocate budgets based on financial data. Three configurations were compared:

\begin{itemize}
    \item RL+DP: Reinforcement Learning with Dirichlet Priors and Bayesian belief updation
    \item RL+DP+Genetic: RL enhanced with a Genetic Optimization module
    \item RL+DP+Genetic+Quantum: An extended configuration incorporating quantum-inspired optimization
\end{itemize}

\noindent The performance of each configuration was evaluated using the following four metrics:

\begin{itemize}
    \item MAE (Mean Absolute Error)
    \item RMSE (Root Mean Square Error)
    \item Cosine Similarity
    \item KL Divergence (Kullback--Leibler Divergence \cite{DBLP:journals/corr/Shlens14c})
\end{itemize}

\begin{table}[htbp]
  \centering
  \tiny
  \resizebox{\columnwidth}{!}{%
  \begin{tabular}{|l|l|l|l|l|}
    \hline
    \textbf{Model} & \textbf{MAE} & \textbf{RMSE} &\textbf{Cosine Similarity} &\textbf{KL Divergence} \\
    \hline
    RL+DP       &0.1047    &0.1044 & 0.8832      &0.0110  \\
    RL+DP+Genetic      & 0.0995   & 0.0861&0.9813   & 0.0099    \\
    RL+DP+Genetic+Quantum     & 0.0229   &0.0283 &0.9990  & 0.0023   \\
   
    \hline
  \end{tabular}%
  }
  \caption{Budget Allocation Results (The best scores have been obtained by setting SEED=60) DP stands for Dirichlet Priors}
  \label{tab:results1}
\end{table}

\begin{figure}[htbp]
  \centering
  \begin{minipage}[b]{0.48\textwidth}
    \centering
    \includegraphics[width=8cm,height=4cm]{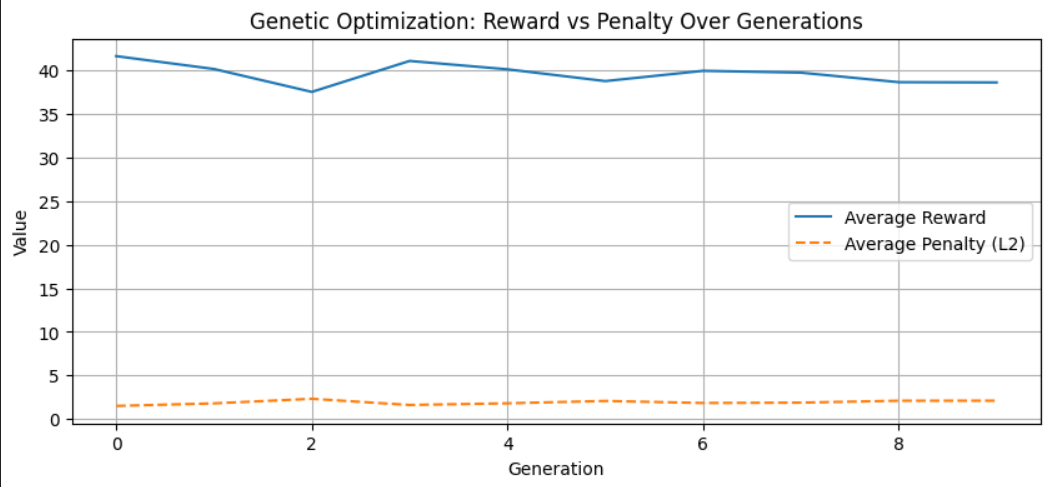}
    \caption{GA Optimization}
    \label{fig:results4}
  \end{minipage}
  \hfill
  \begin{minipage}[b]{0.48\textwidth}
    \centering
    \includegraphics[width=8cm,height=4cm]{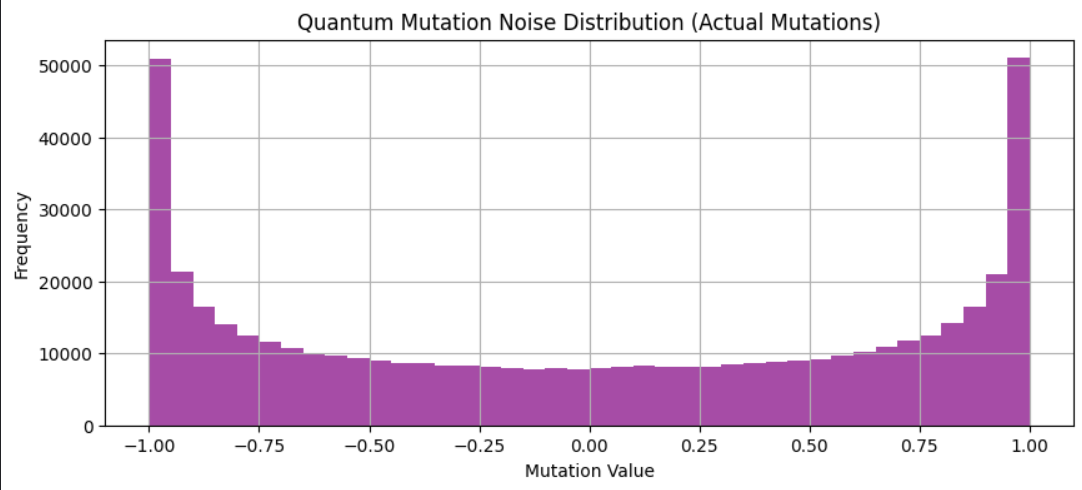}
    \caption{Quantum Mutations}
    \label{fig:results5}
  \end{minipage}
\end{figure}

The predicted allocations for the test set have been shown in  \autoref{fig:results1}
\noindent The genetic optimization reward curve in \autoref{fig:results4} demonstrates the effectiveness of evolutionary fine-tuning, where average policy performance improves steadily across generations, indicating successful selection and recombination of high-performing individuals. While minor fluctuations are observed due to the stochastic nature of the environment and exploration dynamics, the overall upward trend highlights convergence toward more optimal budget allocation strategies. Complementing this, the quantum mutation noise distribution reveals a balanced spread of parameter perturbations around zero, introduced via quantum-inspired rotations. \autoref{fig:results5}

\section{Conclusion}
This study introduces a practical and modular hybrid framework for intelligent budget allocation, combining TD3-based deep reinforcement learning, probabilistic modeling via Dirichlet priors, and a quantum-inspired genetic optimization layer. Using real-world quarterly financial data from Apple Inc.\ (2009--2025), the model learns to dynamically allocate R\&D and SG\&A spending to optimize long-term net income under realistic constraints.

The custom Gym environment simulates economic decision-making under uncertainty, incorporating Bayesian updating of budget priorities and a reward function that penalizes deviation from historical allocations via L2 loss. This dual-objective reward---balancing profitability and allocation realism---guides the learning process toward both strategic and interpretable policies.

Training with Stable-Baselines3's TD3 agent, enhanced by NormalActionNoise and a debug callback for reward tracing, yielded stable convergence. Post-training, we applied a quantum-inspired genetic algorithm using PennyLane, which introduced PauliZ expectation-based perturbations to fine-tune the agent’s policy weights. This innovative mutation strategy enriches exploration and helped identify parameter configurations that further boosted test performance.

Evaluation on held-out test data shows strong alignment between predicted and actual budget ratios, with mean absolute error (MAE) and root mean squared error (RMSE) values remaining low for both R\&D and SG\&A allocations. Visualization plots confirm the model’s ability to track temporal shifts in financial strategy, suggesting high fidelity in modeling corporate spending behavior.

Looking forward, this framework can be extended by:
\begin{itemize}
    \item Enriching the state space with macroeconomic indicators (e.g., interest rates, GDP trends),
    \item Introducing transaction frictions or regulatory limits as constraints in the environment,
    \item Testing quantum mutation on real hardware via Qiskit or Rigetti Forest,
    \item And evolving into a multi-agent system where departments act as competing/cooperative agents under a shared fiscal policy.
\end{itemize}

In conclusion, this research offers a strategic and data-driven framework that enables organizations to make more informed and optimized financial budget allocation decisions. By combining deep reinforcement learning, probabilistic modeling through Dirichlet priors, and quantum-inspired genetic optimization, the approach moves beyond static, heuristic methods to deliver adaptive, forward-looking policies. The model’s ability to learn from real-world financial data and simulate multi-quarter decision-making ensures that budget allocations—particularly between R\&D and SG\&A—are aligned with long-term profitability and operational goals. With its dual-objective reward design, the system not only seeks maximum return but also adheres to historically realistic and interpretable allocation patterns, increasing stakeholder trust and decision transparency. Ultimately, this research empowers firms—especially those in fast-moving, innovation-driven sectors—to optimize financial planning under uncertainty, enhancing their ability to generate sustained and guaranteed returns over time.
\section*{Financial Disclosure}
The authors declare the financial support received for this work. 

\section*{Conflicts of Interest}
None reported. 

\bibliographystyle{IEEEtran} 
\bibliography{references}

\begin{thebibliography}{10}
\providecommand{\url}[1]{#1}
\csname url@samestyle\endcsname
\providecommand{\newblock}{\relax}
\providecommand{\bibinfo}[2]{#2}
\providecommand{\BIBentrySTDinterwordspacing}{\spaceskip=0pt\relax}
\providecommand{\BIBentryALTinterwordstretchfactor}{4}
\providecommand{\BIBentryALTinterwordspacing}{\spaceskip=\fontdimen2\font plus
\BIBentryALTinterwordstretchfactor\fontdimen3\font minus \fontdimen4\font\relax}
\providecommand{\BIBforeignlanguage}[2]{{%
\expandafter\ifx\csname l@#1\endcsname\relax
\typeout{** WARNING: IEEEtran.bst: No hyphenation pattern has been}%
\typeout{** loaded for the language `#1'. Using the pattern for}%
\typeout{** the default language instead.}%
\else
\language=\csname l@#1\endcsname
\fi
#2}}
\providecommand{\BIBdecl}{\relax}
\BIBdecl

\bibitem{10.1371/journal.pone.0288627}
\BIBentryALTinterwordspacing
A.~Yeganeh and S.~C. Shongwe, ``A novel application of statistical process control charts in financial market surveillance with the idea of profile monitoring,'' \emph{PLOS ONE}, vol.~18, no.~7, pp. 1--26, 07 2023. [Online]. Available: \url{https://doi.org/10.1371/journal.pone.0288627}
\BIBentrySTDinterwordspacing

\bibitem{10.1093/icc/11.3.529}
\BIBentryALTinterwordspacing
H.~Chesbrough and R.~S. Rosenbloom, ``The role of the business model in capturing value from innovation: evidence from xerox corporation's technology spin‐off companies,'' \emph{Industrial and Corporate Change}, vol.~11, no.~3, pp. 529--555, 06 2002. [Online]. Available: \url{https://doi.org/10.1093/icc/11.3.529}
\BIBentrySTDinterwordspacing

\bibitem{CAO2025247}
\BIBentryALTinterwordspacing
Z.~Cao, H.~Wang, E.~P. Chew, H.~Li, and K.~C. Tan, ``A budget-adaptive allocation rule for optimal computing budget allocation,'' \emph{European Journal of Operational Research}, vol. 325, no.~2, pp. 247--260, 2025. [Online]. Available: \url{https://www.sciencedirect.com/science/article/pii/S0377221725002656}
\BIBentrySTDinterwordspacing

\bibitem{10895709}
S.~Trivedi, S.~Roy, R.~Das, D.~Mukherjee, S.~Mahapatra, and P.~Nandy, ``A dual-model solution for business reasoning with large language models,'' in \emph{2024 IEEE Pune Section International Conference (PuneCon)}, 2024, pp. 1--6.

\bibitem{ghasemi2024introductionreinforcementlearning}
\BIBentryALTinterwordspacing
M.~Ghasemi and D.~Ebrahimi, ``Introduction to reinforcement learning,'' 2024. [Online]. Available: \url{https://arxiv.org/abs/2408.07712}
\BIBentrySTDinterwordspacing

\bibitem{han2020actorcriticreinforcementlearningcontrol}
\BIBentryALTinterwordspacing
M.~Han, L.~Zhang, J.~Wang, and W.~Pan, ``Actor-critic reinforcement learning for control with stability guarantee,'' 2020. [Online]. Available: \url{https://arxiv.org/abs/2004.14288}
\BIBentrySTDinterwordspacing

\bibitem{712192}
R.~Sutton and A.~Barto, ``Reinforcement learning: An introduction,'' \emph{IEEE Transactions on Neural Networks}, vol.~9, no.~5, pp. 1054--1054, 1998.

\bibitem{bai2024reviewreinforcementlearningfinancial}
\BIBentryALTinterwordspacing
Y.~Bai, Y.~Gao, R.~Wan, S.~Zhang, and R.~Song, ``A review of reinforcement learning in financial applications,'' 2024. [Online]. Available: \url{https://arxiv.org/abs/2411.12746}
\BIBentrySTDinterwordspacing

\bibitem{macrotrends_aapl}
M.~LLC, ``Apple income statement 2010--2024 | aapl,'' \url{https://www.macrotrends.net/stocks/charts/AAPL/apple/income-statement?freq=A}, 2024, accessed: July 22, 2025.

\bibitem{10823421}
R.~Das, N.~Dutta, D.~Mukherjee, S.~Mahapatra, I.~Mitra, and P.~Nandy, ``Harnessing the power of quantum feature representation for leveraging fused descriptor definition in breast cancer diagnosis,'' in \emph{2024 International Conference on Electrical, Communication and Computer Engineering (ICECCE)}, 2024, pp. 1--6.

\bibitem{mnih2013playingatarideepreinforcement}
\BIBentryALTinterwordspacing
V.~Mnih, K.~Kavukcuoglu, D.~Silver, A.~Graves, I.~Antonoglou, D.~Wierstra, and M.~Riedmiller, ``Playing atari with deep reinforcement learning,'' 2013. [Online]. Available: \url{https://arxiv.org/abs/1312.5602}
\BIBentrySTDinterwordspacing

\bibitem{a16010023}
\BIBentryALTinterwordspacing
M.~Tran, D.~Pham-Hi, and M.~Bui, ``Optimizing automated trading systems with deep reinforcement learning,'' \emph{Algorithms}, vol.~16, no.~1, 2023. [Online]. Available: \url{https://www.mdpi.com/1999-4893/16/1/23}
\BIBentrySTDinterwordspacing

\bibitem{nakayama2023causalinferenceinvestmentconstraints}
\BIBentryALTinterwordspacing
Y.~Nakayama and T.~Sawaki, ``Causal inference on investment constraints and non-stationarity in dynamic portfolio optimization through reinforcement learning,'' 2023. [Online]. Available: \url{https://arxiv.org/abs/2311.04946}
\BIBentrySTDinterwordspacing

\bibitem{huang2025deepreinforcementlearningframework}
\BIBentryALTinterwordspacing
G.~Huang, X.~Zhou, and Q.~Song, ``A deep reinforcement learning framework for dynamic portfolio optimization: Evidence from china's stock market,'' 2025. [Online]. Available: \url{https://arxiv.org/abs/2412.18563}
\BIBentrySTDinterwordspacing

\bibitem{a17120570}
\BIBentryALTinterwordspacing
F.~Espiga-Fern\'{a}ndez, A.~Garc\'{i}a-S\'{a}nchez, and J.~Ordieres-Mer\'{e}, ``A systematic approach to portfolio optimization: A comparative study of reinforcement learning agents, market signals, and investment horizons,'' \emph{Algorithms}, vol.~17, no.~12, 2024. [Online]. Available: \url{https://www.mdpi.com/1999-4893/17/12/570}
\BIBentrySTDinterwordspacing

\bibitem{JIANG2024101016}
\BIBentryALTinterwordspacing
Y.~Jiang, J.~Olmo, and M.~Atwi, ``Deep reinforcement learning for portfolio selection,'' \emph{Global Finance Journal}, vol.~62, p. 101016, 2024. [Online]. Available: \url{https://www.sciencedirect.com/science/article/pii/S1044028324000887}
\BIBentrySTDinterwordspacing

\bibitem{Vlassis2012}
\BIBentryALTinterwordspacing
N.~Vlassis, M.~Ghavamzadeh, S.~Mannor, and P.~Poupart, \emph{Bayesian Reinforcement Learning}.\hskip 1em plus 0.5em minus 0.4em\relax Berlin, Heidelberg: Springer Berlin Heidelberg, 2012, pp. 359--386. [Online]. Available: \url{https://doi.org/10.1007/978-3-642-27645-3_11}
\BIBentrySTDinterwordspacing

\bibitem{KANG2024107924}
\BIBentryALTinterwordspacing
P.~Kang, P.~N. Tobler, and P.~Dayan, ``Bayesian reinforcement learning: A basic overview,'' \emph{Neurobiology of Learning and Memory}, vol. 211, p. 107924, 2024. [Online]. Available: \url{https://www.sciencedirect.com/science/article/pii/S1074742724000352}
\BIBentrySTDinterwordspacing

\bibitem{roy2025generalizedbayesiandeepreinforcement}
\BIBentryALTinterwordspacing
S.~S. Roy, R.~G. Everitt, C.~P. Robert, and R.~Dutta, ``Generalized bayesian deep reinforcement learning,'' 2025. [Online]. Available: \url{https://arxiv.org/abs/2412.11743}
\BIBentrySTDinterwordspacing

\bibitem{Ghavamzadeh_2015}
\BIBentryALTinterwordspacing
M.~Ghavamzadeh, S.~Mannor, J.~Pineau, and A.~Tamar, ``Convex optimization: Algorithms and complexity,'' \emph{Foundations and Trends® in Machine Learning}, vol.~8, no. 5–6, p. 359–483, 2015. [Online]. Available: \url{http://dx.doi.org/10.1561/2200000049}
\BIBentrySTDinterwordspacing

\bibitem{preil2023geneticmultiarmedbanditsreinforcement}
\BIBentryALTinterwordspacing
D.~Preil and M.~Krapp, ``Genetic multi-armed bandits: a reinforcement learning approach for discrete optimization via simulation,'' 2023. [Online]. Available: \url{https://arxiv.org/abs/2302.07695}
\BIBentrySTDinterwordspacing

\bibitem{ALLEN1999245}
\BIBentryALTinterwordspacing
F.~Allen and R.~Karjalainen, ``Using genetic algorithms to find technical trading rules,'' \emph{Journal of Financial Economics}, vol.~51, no.~2, pp. 245--271, 1999. [Online]. Available: \url{https://www.sciencedirect.com/science/article/pii/S0304405X9800052X}
\BIBentrySTDinterwordspacing

\bibitem{narayanan1996quantum}
A.~Narayanan and M.~Moore, ``Quantum-inspired genetic algorithm,'' in \emph{Proceedings of the IEEE International Conference on Evolutionary Computation (ICEC)}, 1996, pp. 61--66.

\bibitem{article}
M.~Naik and A.~Albuquerque, ``Hybrid optimization search-based ensemble model for portfolio optimization and return prediction in business investment,'' \emph{Progress in Artificial Intelligence}, vol.~11, 08 2022.

\bibitem{brockman2016openaigym}
\BIBentryALTinterwordspacing
G.~Brockman, V.~Cheung, L.~Pettersson, J.~Schneider, J.~Schulman, J.~Tang, and W.~Zaremba, ``Openai gym,'' 2016. [Online]. Available: \url{https://arxiv.org/abs/1606.01540}
\BIBentrySTDinterwordspacing

\bibitem{fujimoto2018addressingfunctionapproximationerror}
\BIBentryALTinterwordspacing
S.~Fujimoto, H.~van Hoof, and D.~Meger, ``Addressing function approximation error in actor-critic methods,'' 2018. [Online]. Available: \url{https://arxiv.org/abs/1802.09477}
\BIBentrySTDinterwordspacing

\bibitem{pmlr-v9-glorot10a}
\BIBentryALTinterwordspacing
X.~Glorot and Y.~Bengio, ``Understanding the difficulty of training deep feedforward neural networks,'' in \emph{Proceedings of the Thirteenth International Conference on Artificial Intelligence and Statistics}, ser. Proceedings of Machine Learning Research, Y.~W. Teh and M.~Titterington, Eds., vol.~9.\hskip 1em plus 0.5em minus 0.4em\relax Chia Laguna Resort, Sardinia, Italy: PMLR, 13--15 May 2010, pp. 249--256. [Online]. Available: \url{https://proceedings.mlr.press/v9/glorot10a.html}
\BIBentrySTDinterwordspacing

\bibitem{bergholm2022pennylaneautomaticdifferentiationhybrid}
\BIBentryALTinterwordspacing
V.~B. et~al, ``Pennylane: Automatic differentiation of hybrid quantum-classical computations,'' 2022. [Online]. Available: \url{https://arxiv.org/abs/1811.04968}
\BIBentrySTDinterwordspacing

\bibitem{paszke2019pytorchimperativestylehighperformance}
\BIBentryALTinterwordspacing
A.~Paszke, S.~Gross, F.~Massa, A.~Lerer, J.~Bradbury, G.~Chanan, T.~Killeen, Z.~Lin, N.~Gimelshein, L.~Antiga, A.~Desmaison, A.~Köpf, E.~Yang, Z.~DeVito, M.~Raison, A.~Tejani, S.~Chilamkurthy, B.~Steiner, L.~Fang, J.~Bai, and S.~Chintala, ``Pytorch: An imperative style, high-performance deep learning library,'' 2019. [Online]. Available: \url{https://arxiv.org/abs/1912.01703}
\BIBentrySTDinterwordspacing

\bibitem{JMLR:v22:20-1364}
\BIBentryALTinterwordspacing
A.~Raffin, A.~Hill, A.~Gleave, A.~Kanervisto, M.~Ernestus, and N.~Dormann, ``Stable-baselines3: Reliable reinforcement learning implementations,'' \emph{Journal of Machine Learning Research}, vol.~22, no. 268, pp. 1--8, 2021. [Online]. Available: \url{http://jmlr.org/papers/v22/20-1364.html}
\BIBentrySTDinterwordspacing

\bibitem{DBLP:journals/corr/Shlens14c}
\BIBentryALTinterwordspacing
J.~Shlens, ``Notes on kullback-leibler divergence and likelihood,'' \emph{CoRR}, vol. abs/1404.2000, 2014. [Online]. Available: \url{http://arxiv.org/abs/1404.2000}
\BIBentrySTDinterwordspacing

\end{thebibliography}

\end{document}